# Explainable Multi-class Classification of Medical Data

YuanZheng Hu
EECS, University of Ottawa

Marina Sokolova
IBDA@Dalhousie University and
University of Ottawa

*bhu078@uottawa.ca*

*sokolova@uottawa.ca*

**Abstract** Machine Learning applications have brought new insights into a secondary analysis of medical data. Machine Learning helps to develop new drugs, define populations susceptible to certain illnesses, identify predictors of many common diseases. At the same time, Machine Learning results depend on convolution of many factors, including feature selection, class (im)balance, algorithm preference, and performance metrics.

In this paper, we present explainable multi-class classification of a large medical data set. We in details discuss knowledge-based feature engineering, data set balancing, best model selection, and parameter tuning. Six algorithms are used in this study: Support Vector Machine (SVM), Naïve Bayes, Gradient Boosting, Decision Trees, Random Forest, and Logistic Regression. Our empirical evaluation is done on the UCI *Diabetes 130-US hospitals for years 1999-2008 dataset*, with the task to classify patient hospital re-admission stay into three classes: 0 days, <30 days, or > 30 days.

Our results show that using 23 medication features in learning experiments improves Recall of five out of the six applied learning algorithms. This is a new result that expands the previous studies conducted on the same data. Gradient Boosting and Random Forest outperformed other algorithms in terms of the three-class classification Accuracy.

## 1. Introduction

With the advent and development of state-of-art learning algorithms, Machine Learning has been applied in such diverse domains as finance, medicine, and arts. Auxiliary of Machine Learning has become one of the efficient ways in the medical area for new drug discovery, patient diagnosis, and disease analytics (Pesaranghader et al, 2016). At the same time, efficiency of Machine Learning applications depends on convolution of many factors, where feature selection, algorithm preference, performance evaluation, to name a few factors, contribute to the results.

In this work, we present a comprehensive study of a Machine Learning application in the medical domain. Our task is to perform multi-class classification of a large data set and investigate factors that affect classification results. The dataset we are going to use in this report is provided by UCI Machine Learning repository, named *"Diabetes 130-US hospitals for years 1999-2008 dataset"*, the dataset is originally authored by Strack et.al. (2014). We classify patient hospital re-admission stay into three classes: 0 days, <30 days, or > 30 days. The problem is to find Machine Learning technique that best classifies whether a patient will be readmitted within 30 days, beyond 30 days or they will not be readmitted.

Our work starts with employing the medical domain knowledge to perform feature engineering, followed by construction of a balanced data set. We in-depth investigate performance of six algorithms: Support Vector Machine (SVM), Naïve Bayes, Gradient Boosting, Decision Trees, Random Forest, and Logistic Regression. After a hyperparameter tuning for the six algorithms, we use statistical analysis to pick the best two model for further evaluation. Our work contributes to building Explainable AI (Holzinger et al, 2017): we use medical evidence in feature engineering, in details discuss model functionality, and report procedures of hyperparameter tuning. We compare the algorithm performance based on the obtained Accuracy, and Macro Recall, Precision and F-1 score (Sokolova & Lapalme, 2009).

Our results show that Gradient Boosting and Random Forest outperform other algorithms in terms of Accuracy of the three-class classification. Our results show that using 23 medication features in learning experiments improves Recall of five out of the six applied learning algorithms (Naïve Bayes being the solo exception). This result expands the previous studies conducted on the same data.

## 2. Related Work

Medical domain has a long tradition of being a major application field for Machine Learning (Shawe-Taylor & Christianini, 2004). Among those applications, studies of Diabetes play a prominent role due to severity of the disease and availability of publicly accessible data. Centers for Disease Control and Prevention (CDC, 2017) estimated Diabetes as a disease that severely affects the USA population: 9.4% of the Americans have diabetes while 26% of Americans have prediabetes[1]. World Health Organization (WHO) showed that world-wide 1 out of 11 people has Diabetes.[2] The disease complications can lead to a stroke, blindness, heart attack, kidney failure, and limb amputation. UCI data repository[3] lists several Diabetes sets available for Machine Learning research. Among them, the *"Diabetes 130-US hospitals for years 1999-2008 dataset"* is the largest and most comprehensive set (Strack et.al, 2014). We use this data set in our empirical study. In this section we review the previous work conducted on the same data (the Diabetes data hereafter).

The Diabetes data has been used for binary classification of the patient's readmission, resulting in classification of two classes: *< 30* days or *>30* days. For example, Ravindra et al. (2016) compared SVM, Logistic Regression, Decision Trees, Random Forest, and Genialized Boosted Modeling with the standardized readmission tool (LACE). They concluded that Machine Learning algorithms are promising when compared with LACE. Bhuvan et.al (2016) performed an additional cost analysis on patient's readmission and the binary prediction on if the patient will be readmitted in 30 days or would not. They concluded that Random Forest and Neural Network are the best model with an area of 0.65 and 0.654 by using under Precision-Recall Curve. The same binary task was tackled by Hammoudeh et.al. (2018) . The authors used CNN and achieved an accuracy of 92%. Unfortunately, the authors did not give the details of the model architecture; we only know that an additional early stopping technique was applied. Li et.al (2018) achieved accuracy of 88% by using Collaborative Filtering- enhanced Deep Learning (CFDL), also for the binary classification of the readmission days. Reid (2019) achieved 86% accuracy of binary classification on readmission days by using Boosted Decision Tree.

A few patient-centric studies were conducted on the same data set. Graham et.al (2019) applied Logistic Regression and SVM to identify 11 major variables that may have impact on the readmission days. The resulting variables are age, race, discharge disposition id, insulin, readmitted, admission type id, admission source id, time hospital, number diagnoses, ismale, and being on diabetes medications or not. By using Deep Neural Network, Avram et.al. (2020) has correctly identified 82 percent of the patient with diabetes. Improvement of patients' privacy protection has been the goal of multi-label classification of demographic features (age, race, gender) in Cotha & Sokolova (2015) and Tran & Sokolova (2016). Jafer et al (2017) proposed privacy-enhancing evaluation of Machine Learning results, where the algorithm performance is judged by combination of the classification accuracy and protection of patients' demographic features.

---

[1] https://www.cdc.gov/media/releases/2017/p0718-diabetes-report.html

[2] https://www.who.int/news-room/fact-sheets/detail/diabetes

[3] UCI Machine Learning Repository

We, however, worked on the multi-classification of the patient's readmission, which resulted in classification of three classes: *0, <30, or > 30* days. This is a novel task that provides more insights into patient readmission prospects than the binary classification tasks mentioned above. The results of the three-class classification can lead to finer-grained patient-centric and privacy protection analysis of the data.

## 3. The Data Processing

For the empirical evidence, we use the Diabetes dataset, a large publicly available set. The set and its features were introduced by Strack et.al. (2014) and deposited to the UCI Machine Learning Repository [4]. The dataset has 100000 instances and 55 features; we provide the detailed characteristics below

### 3.1 Feature type categorization

Our first step is to categorize and analyze the data in detail, convert and transform the data type into the best type that fits the problem of multi-classification. This means that we need to convert those data features based on their relevancy regarding to the readmission days and their quantitative meaning. We first encode the data features based on their usage in diabetes treatment and their type in the dataset. Table 1 reports the features and their types. Our decision for the type categorization is based on Strack et.al (2014).

| Data types | Feature name |
|---|---|
| **Nominal** | race, gender, admission type, discharge disposition, admission source, medical specialty, change of medications, diabetes medications. |
| **Ordinal** | glucose serum test result, A1c test result, 24 medication features (the features are listed in Appendix 1) |
| **Numerical** | payer code, diagnosis 1, diagnosis 2, diagnosis 3, encounter ID, patient number, weight, time in hospital, number of lab procedures, number of procedures, number of medications, number of outpatient visits, number of emergency visits, number of impatient visits, number of diagnoses. |
| **Interval** | Age |

*Table 1. Feature categorization for the Diabetes data set.*

The *ordinal type* includes glucose serum and A1c test results along with the 24 medication features. Glucose serum or blood sugar level test result has values of ">200", ">300", "Normal" and "None". This can be used an indication of diabetes. According to American Diabetes Association, a healthy people have below 100 mg/dL sugar level, any values above this threshold can indicate diabetic[5]. A1c test result is another type of test for diabetes, it has values of ">7", ">8", "Normal" and "None". Healthy people have should have range below 5.7, any values above this can indicate prediabetic or diabetic. The medication 24 features are names of 24 different medication prescriptions and their dosage during the encounter; the features have values of "up", "down", "steady" and "no", indications of their usage. All the

---

[4] UCI Machine Learning Repository: Diabetes 130-US hospitals for years 1999-2008 Data Set
[5] https://www.diabetes.org/a1c/diagnosis

26 features mentioned above have a different range of values, and each range of values indicates their level of health condition and the medication usage.

For the ***numerical type***, payer code and diagnosis 1,2,3 are numerical in the dataset; they refer to the encoded payment method and diagnosis name. they need to be decoded using the additional csv file provided. Diagnosis 1,2,3 refer to the diagnosis given by the doctor; they have been preprocessed into a numerical indication of diseases. In this step, we only take care of the actual data type in the dataset; their usage in prediction will be further explored on the next step.

For the ***interval type***, by manually inspection we found age feature is been categorized into 10 range value, namely [0-10], [10-20], [20-30], [30-40], [40-50], [50-60], [60-70], [70-80], [80-90], [90-100]. This is an obvious interval type.

The remaining ***nominal type*** features are represented as characters in the dataset. None of the features reflects the level of importance with respect to the prediction of the readmission.

## 3.2 The data set cleaning

We minimize the number of instances with the empty or noisy features. Specifically, we first remove weight, payer code, and medical specialty features that miss 97%, 52%, and 53% of their values (Strack et.al, 2014). After manually inspection, we found that null data are labeled as "?" in the original dataset. To better analyze the data, we will replace all "?" in the dataset with N/A, and we plot the following figure based on the missing values of each feature.

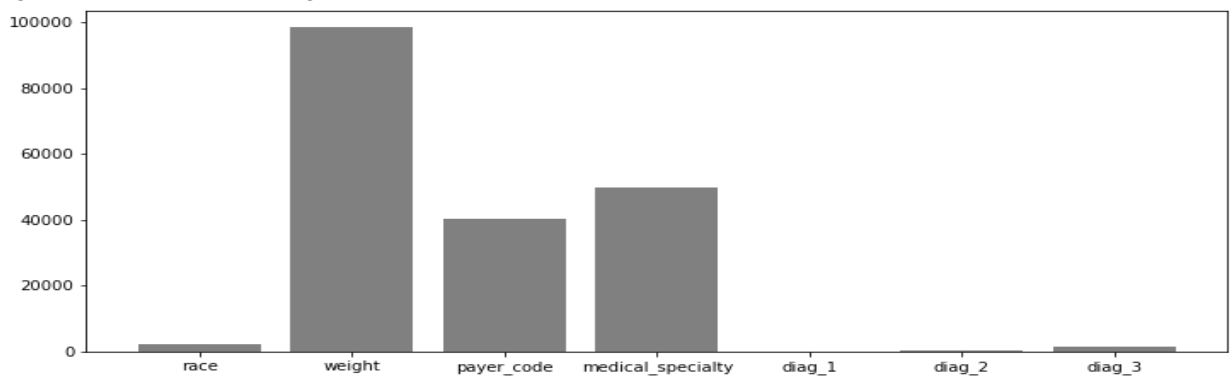

*Figure 1, Instances with empty values in the dataset*

As we observe from Figure 1, the weight, payer code, and medical specialty are indeed highly missing features. For the rest of the features, race, diag_2 and diag_3, also contain the null values. We removed all the instances where missing values appears. Other special indication of noise values will also be removed. In the race feature, we remove all the instances with the value "Unknown/Invalid". The dataset consists of 52337 instances of "No", 34649 instances of ">30" and 11066 instances of "<30".

## 3.3 Feature processing

Our goal is to optimize the features and their values by using information provided by medical experts. This helps to minimize data overbudance, which poses challenges for Machine Learning applications in medical and health care domains (Vellido, 2019).

### 3.3.1 Nominal type processing

For the race, change of medications, diabetes medications, we use binary encoding or dummy encoding that indicates existence of a variable by using numeric 0 or 1. We use it since magnitude of the numerical labels might introduce controversy to the data analysis. For example, if race been encoded by using numerical values, the Machine Learning models will interpret the race as a quantitative feature and leads to unfairness in prediction. For the rest of the nominal data, we will handle them as follows.

For the admission type, we found 8 types by using the ID mapping file with the interpretation of the numerical encoded values in the data. The types are emergency, urgent, elective, newborn, not available, trauma center, and not mapped. Emergency admission shows a critical condition of a patient, thus differs from the other admission types. We reflect on that by reducing the 8 types to 2 types, i.e., emergency and other admissions.

For the discharge disposition, by using the same ID mapping file, we identified 3 types, i.e., home, transfer, and the other discharge. Those values relate to patient's health condition: discharge to home means the patient might be deemed healthy or has mild symptoms compared to others, transfer means the patient might be discharged to another hospital or health care center for further treatment, the other refers to the patient's discharge disposition is not recorded or expired.

For the admission sources, we identify 3 main types: emergency room, transfer, and the other sources. The emergency room is considered as a higher severity compare to transfer and other types, and transfer is considered as chronic diseases or mild symptom, the other type refers to not recorded or special type admissions like born in this hospital.

### 3.3.2 Ordinal type processing

We map the ordinal features according as follows: according to American Diabetes Association, a healthy person's glucose serum should be ranged below 100 mg/dL, and A1c test result should be around 5%. Thus, we take the norm as 100 and 5 respectively for glucose serum and A1c test results. For the rest of the attributes of the feature, we will take the value on the boundary. As for the 24 medication features of medicines, we only encode insulin by using binary encoding since insulin is a common treatment for diabetes. According to (Bhuvan et.al., 2016), only insulin shows a significant variance among the patients, while the other 23 features do not have significant variance. We analyzed the data value for the 24 features including insulin, and found that their values are represented by "Up", "Down","Steady", and "No". Considering the space of our machine, and the memory requirement on deep learning models, we perform binary encoding on insulin only; for the rest of 23 features, we will represent their attributes by 1,-1,0, and -2 respectively. Note that we plan to investigate the 23 medication features and their impact in more details after we build the learning models. Table 2 summarizes the results.

| Feature name | Mapping |
| --- | --- |
| glucose serum test result | None: 0, Norm: 100, >200: 200, >300: 300 |
| A1c test result | None: 0, Norm: 5, >7: 7, >8: 8 |
| 24 medication features | Insulin: binary encoding, 23 features: 1, -1, 0, or -2. |

*Table 2 Ordinal feature processing*

### 3.3.3 Numerical and interval type processing

For the numerical features, we drop two randomly assigned features: encounter ID and patient number since they do not provide any useful information for learning. For the rest of the features, we will keep them intact as shown in Table 3. The numerical features show severity of conditions. The diagnosis features use the ICD-9 code. We use "*ICD-9-CM Official Guidelines for Coding and Reporting*" provided by CDC of the U.S. According to Strack et al (2014), in addition to the nominal diagnosis mentioned above, approx. 17.3% of the instances have diagnoses like mental disorders, nervous system disorder, external injury, etc. We label these 17.3% instances as "*other diagnoses*". After nominal encoding, we will perform binary encoding on diagnoses 1,2, and 3.

| Feature name | Mapping |
|---|---|
| diagnosis1,2,3 | 1. 250-251: diabetes, 390-458, 785: circulatory, 460-519, 786: respiratory, 520-579, 787: digestive, 580-629, 788: genitourinary, 800-999: injury, 710-739: musculoskeletal, 140-239: neoplasms<br>2. Binary encoding |
| Time in hospital | Keep same |
| number of lab procedures | Keep same |
| number of procedures | Keep same |
| number of medications | Keep same |
| number of outpatient visits | Keep same |
| number of emergency visits | Keep same |
| number of impatient visits | Keep same |
| number of diagnosis. | Keep same |

*Table 3 numerical feature processing*

Lastly, for the interval typed feature (i.e., age), we represent each range by its median. For example, [0,10] will be encoded as 5, [10,20] will be encoded as 15.

### 3.4 Balancing the Data set with SMOTE

We focus on Machine Learning abilities to classify the readmission days for this dataset. We perform multi-class classification to predict three different classes for the readmission days, which are *0, <30*, and *> 30 days*. As stated in Sec 3.2, the dataset consists of 52337 instances of "No", 34649 instances of ">30" and 11066 instances of "<30". We use the data oversampled by SMOTE[6]. To preserve the original data as much as possible, we keep all 34649 instances of ">30" and randomly pick the same amount of "No" to under-sample the dataset, then perform oversampling methods on the dataset. Figures 2 and 3 show the results.

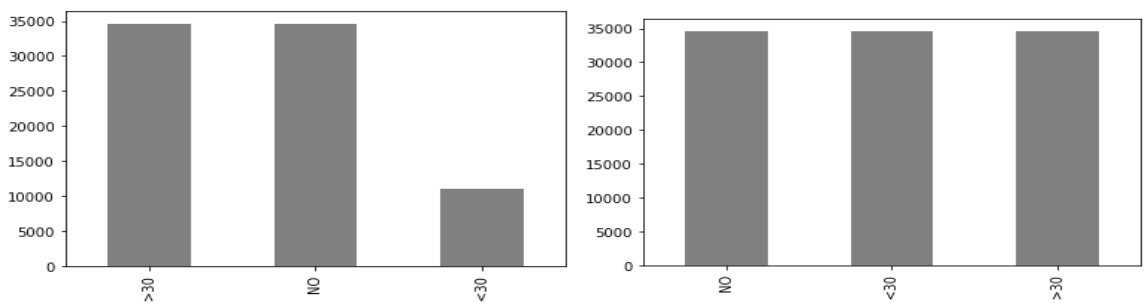

*Figure 2,* imbalanced dataset after under sampling "No"  *Figure 3* balanced dataset after oversampling

## 4. Machine Learning Methods

### 4.1 Preliminary results

We use Naïve Bayes, Gradient Boosting, Random Forest, Decision Trees, SVM, and Logistic Regression to train on our dataset before experimenting on the deep learning models to verify our preprocessed dataset are correctly processed. We the data with the 23 medication features included. The tests in this section are performed on the data split into 80% of training data and 20% of test data, with no cross-validation used.

---
[6] SMOTE - Azure Machine Learning | Microsoft Docs

| Model | Recall macro | Precision macro | F-1 score macro | Accuracy |
|---|---|---|---|---|
| Naive Bayes | 38.9% | 43.84% | 30.32% | 39.10% |
| Gradient Boosting | 64.14% | 67.93% | 65.16% | 64.16% |
| Random Forest | 61.52% | 63.81% | 61.86% | 61.54% |
| Decision Tree | 60.17% | 65.31% | 61.24% | 60.17% |
| Logistic Regression | 45.48% | 45.12% | 44.71% | 45.46% |
| SVM | 53.46% | 52.86% | 52.79% | 53.50% |

*Table 4, Model evaluations, with the 23 features*

Table 4 reports the results of the algorithms obtained with the default settings in Scikit-Learn. Gradient Boosting scored the best accuracy among the six models, whereas Naïve Bayes is the worst among the models. Note that all the algorithms obtained relatively balanced Precision and Recall. This uniform result indicates that there is no data over-fitting. To assess the hypothesis about impact of the 23 medication features, we run the same experiment on the data with those features removed.

| Model | Recall macro | Precision macro | F-1 score macro | Accuracy |
|---|---|---|---|---|
| Naive Bayes | 46.33% | 45.97% | 45.58% | 46.37% |
| Gradient Boosting | 64.25% | 68.32% | 65.33% | 64.27% |
| Random Forest | 61.68% | 64.80% | 62.20% | 61.69% |
| Decision Tree | 59.58% | 64.96% | 60.72% | 59.59% |
| Logistic Regression | 45.29% | 44.93% | 44.54% | 45.28% |
| SVM | 54.29% | 53.74% | 53.72% | 54.33% |

*Table 5 Model evaluation without the 23 features.*

We observed that Naïve Bayes' performance dropped significantly when the 23 features were added. For the other algorithms, we notice that their Recall improved with the 23 features added. This means that the 23 features helped them to correctly identify more patients belonging to each class. Our further experiments run on the data with the 23 features included. We report Micro F-score/Precision/Recall and confusion matrixes in Appendix 2.

### 4.2 Hyperparameter Tuning

We tune the parameters for Gradient Boosting, Decision Trees, Random Forest, Logistic Regression, and SVM. We use a grid search and 5-fold cross-validation for each configuration and take the mean accuracy as the result. In the end, we will select the best configuration for each model.

### 4.2.1 Gradient Boosting

For Gradient Boosting, we pick the parameters based on the following table.

| Parameter name | Value |
|---|---|
| Learning rate | 1, 0.5, 0.1 |
| N_estimator | 50,100,150 |
| Max_depth | 1,2,3,4,5,6 |

*Table 6, tunable parameters for Gradient Boosting*

According to Boehmke et.al. (2020), the principle of Gradient Boosting algorithm is to combine multiple weak learners into a strong learner by using iteration in the form of a decision tree. Considering we have over 100000 instances in this dataset, we pick a slightly large number of trees 150, along with 100 (default) and 50 to test the tendency of accuracy. We will pick 1,2,3 (default),4,5,6 as our maximum depth for the trees. By default, the learning rate is 0.1; since we are using 150 trees as our estimators, we should use a relatively large learning rate to reduce our error, thus we picked 1 and 0.5 to be our learning rate.    We performed 54 tests based on these configurations, each with 5-fold cross-validation, in a total of 270 instances. The total time cost for such training took around 3 hours. We plot our results based on the accuracy and their rank. The accuracy ranged from 55.08% to 64.76%; compared to the accuracy before tuning, 64.16%, there is a slight increase in accuracy. We then explore the spatial distribution for the three parameters.

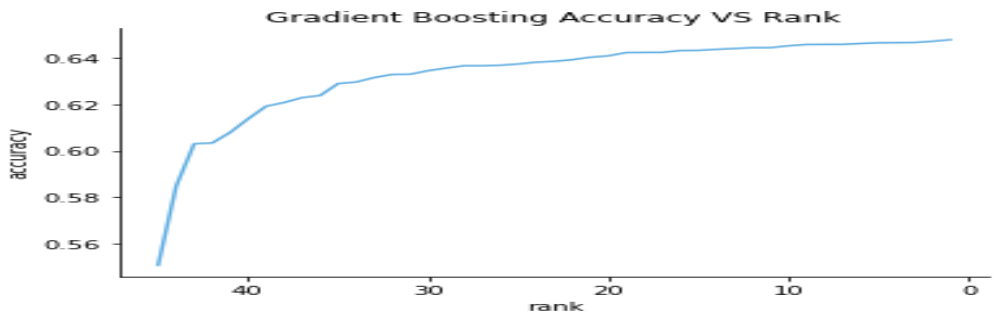

*Figure 4,* Gradient Boosting accuracy based on its parameter tuning

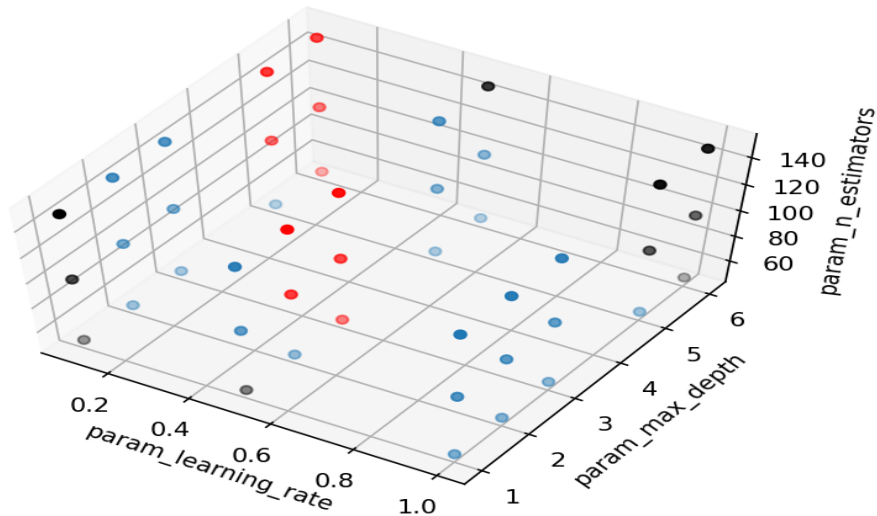

*Figure 5,* spatial distribution for tunable parameters in Gradient Boosting

    The graph above shows the distribution of 54 mean test scores from 270 tests (5 cross-validations each parameter configuration). The red dots represent the top 10 highest test scores, and the black dots represent the bottom 10 lowest test scores. Although it is difficult to generalize on distribution for the top scores, we can hypothesize that the lower scores tend to lie on the edges of the cube. The black dots on the right-hand side indicate that a higher learning rate could result in a worse prediction.

| Parameter | Mean of the top 10 highest accuracy | Std of the top 10 highest accuracy | Mean of the last 10 lowest accuracy | Std of the last 10 lowest accuracy |
|---|---|---|---|---|
| Learning rate | 0.3 | 0.21 | 0.63 | 0.416 |
| N_estimators | 110 | 39.44 | 105 | 43.78 |
| Max depth | 4.1 | 1.66 | 3.8 | 2.44 |

*Table 7, Top ranking parameters and bottom ranking parameters comparison Gradient Boosting*

We also computed the mean and standard deviation for the parameters. As we can see from the Table 8, the number of estimators and max depth does not seem to influence the accuracy of the model, but the learning rate has a significant discrepancy between the mean value of the top 10 highest accuracy instance and last 10 lowest accuracy instances. Lower learning rates tend to have better results than higher ones.

|  | Accuracy | Learning rate | Max depth | N_estimators |
|---|---|---|---|---|
| **Before** | 64.161% | 0.1 | 3 | 100 |
| **After** | 64.769%, | 0.1 | 5 | 150 |

*Table 8 , Parameter tuning result for Gradient Boosting*

### 4.2.2 Random Forest

For Random Forest, we pick the number of estimators, max depth for each tree, min samples to split, and maximum features as parameters to tune. The total training time in this step is around 1-2 hours. Same as before, we are using 5-fold cross-validation with a grid search approach to tuning our parameters. There are 81 cross-validation results and 405 instances in total.

| Parameter | Value |
|---|---|
| N_estimator | 100,200,500 |
| Max_depth | 6,10,20 |
| Min_samples_split | 2,3,4 |
| Max_features | 5,61, auto |

*Table 9, Tunable parameters and their values in Random Forest*

According Kononenko et.al. (2007), Random Forest acts as an ensemble method and gives out the prediction problem to multiple trees and pick the decision with the most vote. Constructing Decision Tree plays an important role in the Random Forest; we use 100 (default), 200, and 500 number of decision trees to enlarge our Random Forest because we have over 100000 instances. The default max depth is expanding until all leaves contain the default minimum number of leaves; since we have limited computing power, we picked 6,10,20 of depth as our max depth. With the same reason for minimum samples to split, we use minimum samples 2 (default), 3, 4 to save computing power. We picked 5, 61, auto (default) as our max features to use in each decision tree to examine the situations without default auto adjustment. The graph below shows the mean test score for each parameter configuration. The test score after parameter tuning ranged from 58.75% to 64.41%, the score was 61.54% before tuning.

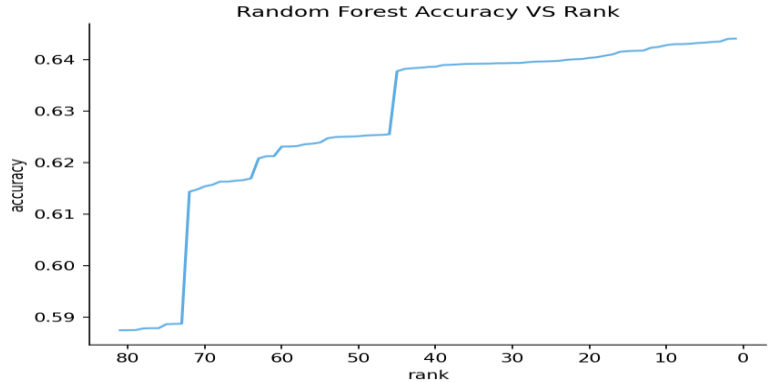

*Figure 2,* Accuracy ranking for tuning parameters in Random Forest

We analyze the parameter more precisely by computing the mean and standard deviation in Table 9. Since max_feature contains an "auto" type, so it can not be calculated for mean and standard deviation, we will use frequency instead. By analyzing the results, we found max_depth is an obvious feature to influence Accuracy of the model. The other two features that have an impact on accuracy are n_estimator and max_features. As we can see from Tables 10 and 11, a larger number of the estimator and lower feature to split will leads to higher accuracy.

| Parameter | Mean (top 10) | Std (top 10) | Mean (last 10) | Std (last 10) | Frequency (top 10) | Frequency (last 10) |
|---|---|---|---|---|---|---|
| max_depth | 20 | 0 | 6 | 0 | N/A | N/A |
| max_features | N/A | N/A | N/A | N/A | 5:5, auto:5 | 61: 9 5: 1 |
| min_samples_split | 3.2 | 0.7 | 2.9 | 0.87 | N/A | N/A |
| n_estimators | 380 | 154 | 250 | 177 | N/A | N/A |

*Table 10, Top ranking parameters and bottom ranking parameters comparison Random Forest*

|  | Accuracy | Max depth | Max feature | Min sample split | N estimator |
|---|---|---|---|---|---|
| Before | 61.54% | 6 | auto | 2 | 100 |
| After | 64.41% | 20 | auto | 4 | 500 |

*Table 11, Parameter tuning result for Random Forest*

### 4.2.3 Decision Tree

For the Decision Tree model, we pick four parameters: max depth for the tree, minimum samples to split a node in the tree, the minimum number of samples required at the leaf level, and the maximum number of features to consider when split. We use 5-fold cross-validation along with parameters' grid search to train our data. Total training time is around 1 hour.

| Parameter name | Value |
|---|---|
| max_depth | 2,5,10, None |
| min_samples_split | 2,3,4,5 |
| min_samples_leaf | 1,2,3 |
| max_features | 10,30,61,auto |

*Table 12, Tunable parameters and their values in Decision Tree*

For all the selected parameters, we increased the values to save the computing power since all selected parameters except for *max_depth* will result in large Decision Tree if their values are small and consequently consuming more computing power. The graph below shows the mean test score for each parameter configuration. The test score after parameter tuning ranged from 41.55% to 60.28%, the score was 60.17% before tuning.

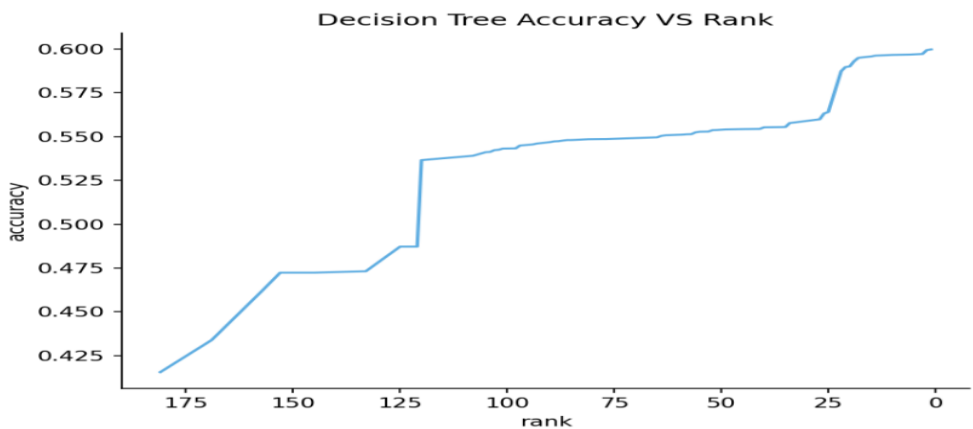

*Figure 3,* Accuracy ranking for tuning parameters in Decision Tree

| Parameter | Mean (top 10) | Std (top 10) | Mean (last 10) | Std (last 10) | Frequency (top 10) | Frequency (last 10) |
|---|---|---|---|---|---|---|
| max_depth | 10 | 0 | 2 | 0 | N/A | N/A |
| max_features | N/A | N/A | N/A | N/A | 61:6 30:5 | auto:10 |
| min_samples_split | 3.3 | 1.16 | 3.6 | 1.17 | N/A | N/A |
| min_samples_leaf | 2.3 | 0.8 | 1.8 | 0.7 | N/A | N/A |

*Table 13, Top ranking parameters and bottom ranking parameters comparison Decision Tree*

From Tables 13 and 14, we conclude is that the maximum depth tends to be higher when accuracy goes up, and maximum depth tends to be lower when accuracy goes down- the same as the graphical analysis.

| | Accuracy | Max depth | Max feature | Min sample split | Min sample leaf |
|---|---|---|---|---|---|
| **Before** | 60.17% | 10 | none | 2 | 100 |
| **After** | 60.28% | 10 | 61 | 3 | 1 |

*Table 14, Parameter tuning result for Decision Tree*

### 4.2.4 Logistic Regression

For Logistic Regression, we tune two parameters: inverse regulation strength and the maximum number of iterations to converge. Our picked the values are shown in the following table. There are 60 cross-validations, 300 instances in total, took around 3 more hours to train.

| Parameter name | Value |
|---|---|
| C | logspace(-4,4,15) |
| max_iter | 5000,10000,20000,30000 |

*Table15, Tunable parameters and their values in Logistic Regression*

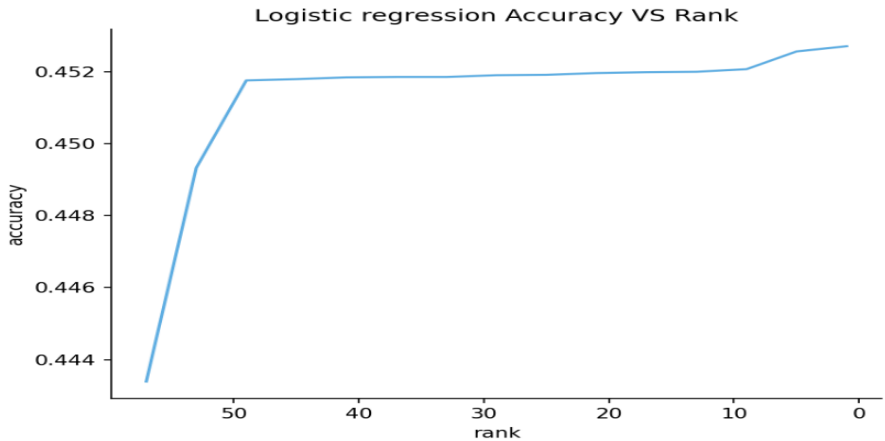

*Figure 4,* Accuracy ranking for tuning parameters in Logistic Regression

The graph above shows the mean test score for each parameter configuration. The test score after parameter tuning ranged from 45.17% to 45.27%; the score was 45.46% before tuning.

| Parameter | Mean (top 10) | Std (top 10) | Mean (last 10) | Std (last 10) |
|---|---|---|---|---|
| C | 0.05 | 0.11 | 2000 | 4216.37 |
| Max iteration | 16500 | 10814.08 | 18000 | 10055.402 |

*Table 16, Top ranking parameters and bottom ranking parameters comparison Logistic Regression*

The graphical distribution is not obvious for Logistic Regression, and instance points with different parameters are visually overlapped, and hard to conclude anything. However, we see lower C values lead to a higher accuracy.

|  | Accuracy | C | Max iteration |
|---|---|---|---|
| **Before** | 45.46% | 1.0 | 100 |
| **After** | 45.27% | 0.05 | 30000 |

*Table 17, Parameter tuning result for Logistic Regression*

### 4.2.5 Support Vector Machine (SVM)

We only select penalty of error. We ran the test with 80% of training data and 20% of test data.

| Parameter name | Value |
|---|---|
| C | 0.1, 1, 10, 100 |

*Table 18, tunable parameters and their values in SVM*

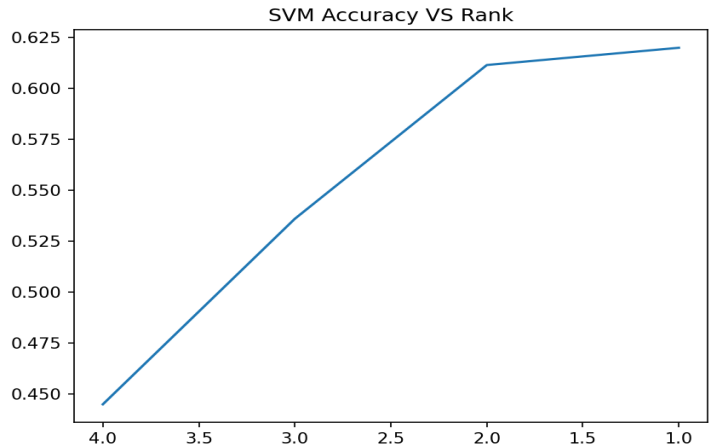

*Figure 5,* accuracy for tuning parameters in SVM

As shown in the graph, the value of C has a critical impact on the accuracy, However, due to the training time constraint, we could not test more values of C.

|  | **Accuracy** | **C** |
|---|---|---|
| **Before** | 53.51% | 1.0 |
| **After** | 61.99% | 100 |

*Table 19, Parameter tuning result for SVM*

We summarize the results in Table 20. It shows the algorithms' Accuracy before and after parameter tuning. Since the dataset is considered as a large dataset, thus, any accuracy discrepancy larger than 1% will be considered as significant.

| **Model** | **Before** | **After** | **Result** |
|---|---|---|---|
| Gradient Boosting | 64.161% | 64.769%, | Insignificant |
| Random Forest | 61.54% | 64.41% | Significant |
| Decision Tree | 60.17% | 60.28% | Insignificant |
| Logistic Regression | 45.46% | 45.27% | Insignificant |
| SVM | 53.51% | 61.99% | Significant |
| Naïve Bayes | 39.10% | N/A | N/A |

*Table 20, Model Accuracy in respect to parameter tuning*

We see that Gradient Boosting and Random Forest scored the highest accuracy among the six models. Thus, we will perform statistical analysis on these two models, to further explore the best Machine Learning model.

# 5. Statistical analysis

## 5.1 The sign test

We analyze the results obtained by Gradient Boosting and Random Forest since those models outperformed Logistic Regression and Decision Tree in terms of accuracy and outperformed SVM in terms of the training time. The parameter configuration for Gradient Boosting and the Random Forest are the best parameters when we performed the hyperparameter tuning: Gradient Boosting: Learning rate = 0.1, max depth = 5, n_estimator = 150; Random Forest: Max depth = 20, min sample split =4, n_estimator = 500. We use the sign test to decide which model performs better.

We propose a null hypothesis that Gradient Boosting and Random Forest perform similarly on the dataset for level alpha equals 0.05.

| Fold # | Gradient Boosting | | | | Random Forest | | | |
|---|---|---|---|---|---|---|---|---|
| | Accuracy | Recall | Precision | f-1 score | Accuracy | Recall | Precision | f-1 score |
| 1 | 64.44 | 64.44 | 68.73 | 65.61 | 64.49 | 64.49 | 69.02 | 65.71 |
| 2 | 64.55 | 64.55 | 69.03 | 65.70 | 64.19 | 64.19 | 69.10 | 65.43 |
| 3 | 64.54 | 64.54 | 69.00 | 65.72 | 64.57 | 64.57 | 69.29 | 65.80 |
| 4 | 64.28 | 64.28 | 68.93 | 65.43 | 63.92 | 63.92 | 68.93 | 65.15 |
| 5 | 65.08 | 65.08 | 69.47 | 66.19 | 65.12 | 65.12 | 69.85 | 66.30 |
| 6 | 65.42 | 65.42 | 69.70 | 66.52 | 65.27 | 65.27 | 69.79 | 66.43 |
| 7 | 64.51 | 64.51 | 69.08 | 65.67 | 64.78 | 64.78 | 69.57 | 65.96 |
| 8 | 65.16 | 65.16 | 69.53 | 66.24 | 64.94 | 64.94 | 69.70 | 66.10 |
| 9 | 64.62 | 64.62 | 69.02 | 65.73 | 64.37 | 64.37 | 69.16 | 65.57 |
| 10 | 64.87 | 64.87 | 69.39 | 66.00 | 65.14 | 65.14 | 69.88 | 66.31 |

*Table 21, Accuracy comparison of Gradient Boosting and Random Forest*

However, from the results obtained on 10 folds (Table 18), we found that GB vs RF, n(f1) = 6 and n(f2) = 4, with w (0.05) = 8, we cannot reject the null hypothesis. Thus, we conclude that Gradient Boosting and Random Forest performed similarly on the dataset.

## 5.2 Boxplot comparison of hyperparameters

We use boxplots to analyze the parameters of Gradient Boosting and Random Forest during parameter tuning. The accuracy shown in the boxplot is the mean accuracy of 5-fold cross-validation.

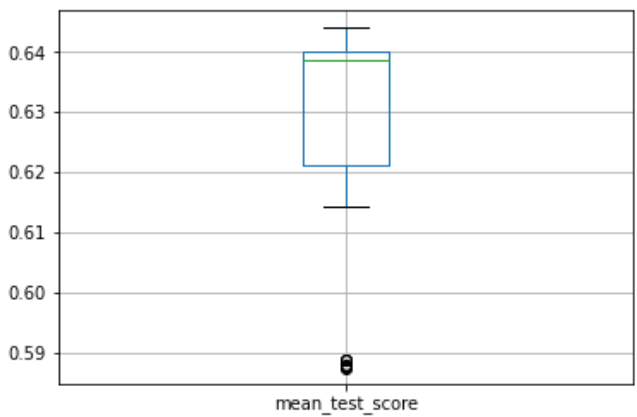

*Figure 6,* Boxplot for Random Forest during parameter tuning

For Random Forest we chose *n_estimator, max_depth, min_samples_split,* and *max_feature* to tune on; for each parameter, we pick a small value, a medium value, and a relatively large value to train on (compare to the default value provided by the library). So, we wish to find a general trend of accuracy.

According to this plot, we found there are 9 outliers reach approx. Accuracy of 59%. By analyzing the outliers, we can find that outliers are having the same low max depth and high max feature; this conclusion is the same as the one we found in the previous section when analyzing the mean value of parameters. In addition to this, we also found that there are in total 82 parameter tuning tests, the combinations of the max depth of 6 and max feature of 61 are 9 in total, which happened to be the entire member outliers. In general, we have the following conclusion.

| Max depth | Max feature | Performance |
|---|---|---|
| Low | High | Bad |
| Low | Low | Average |
| High | Low | Average |

*Table 22, Performance correspondence for Random Forest*

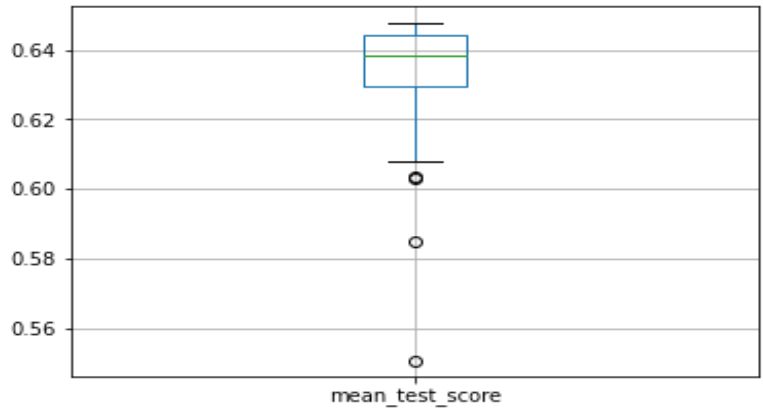

*Figure 7,* Boxplot for Gradient Boosting during parameter tuning

The figure above shows the accuracy boxplot for Gradient Boosting during parameter tuning, we picked *learning_rate, n_estimator, max_depth* as parameter to tune on. As in the case of Random Forest, we choose the values around the default value provided by the Scikit-Learn library to train.

We will also analyze the outliers for Gradient Boosting, there are four outliers, and their parameters are shown in the table below.

| Accuracy | Learning rate | Max depth | n estimator |
|---|---|---|---|
| 60.33 | 1 | 6 | 150 |
| 60.30 | 0.1 | 1 | 150 |
| 58.48 | 0.1 | 1 | 100 |
| 55.08 | 0.1 | 1 | 50 |

*Table 23, Outliers and their parameters*

From the table above, we observed that the combination of learning rate 0.1 and max depth 1 performs badly among 46 tuning tests. In particular, the average accuracy lies around 64%, but the outliers with learning rate 0.1 and max depth 1 have accuracy below 60%. Besides, there are only three combinations of learning rate 0.1 and max depth 1, and they all belong to lower outliers. However, in the section of Gradient Boosting, by analyzing the mean we conclude that a lower learning rate will lead to better accuracy, here we found that the worst three tests have a lower learning rate. Thus, we can add to the conclusion that the combination of low learning rate and low max depth will lead to bad results regardless of the low learning rate value.

However, if we compare the *n_estimator* vertically, we found that the value of the n estimator could positively impact the accuracy regardless of the bad combination of learning rate and max depth. Given that the increase of estimator will increase accuracy also. But still, they all lie below the minimum boundary (Q1 – 1.5IQR).

# Conclusions and Future Work

In this work, we have reported an explainable multi-class classification of a large medical data set (i.e., *Diabetes 130-US hospitals for years 1999-2008 dataset*). We first have explored advantage of feature engineering. We have categorized the features into the nominal, ordinal, interval, and numerical types and cleaned the empty and noisy data. We then preprocessed the data by mapping the encoded ICD-9 code by using the guideline provided by CDC; we transformed and encoded the pre-encoded features (e.g., admission type and discharge deposition) using binary encoding. We also investigated features closely related to diabetes (e.g., HbA1c test results and insulin dosage). We ended feature construction by designing a detailed feature encoding plan for each feature in the dataset. We then applied an oversampling method SMOTE to balance our dataset. This helped to prevent the model overfitting to the largest class.

We applied six Machine Learning models - Gradient Boosting, Decision Trees, Random Forest, Logistic Regression, and SVM - to conduct a multi-class classification on readmission days 0 days, <30 days, or > 30 days. To the best of our knowledge, this task has not been attended in previous studies. We in details discussed the model functionality and reported procedures of hyperparameter tuning.

Our results show that Gradient Boosting and Random Forest are the best among the Machine Learning models in terms of accuracy. They achieved a mean accuracy of 64.769% and 64.41% respectively under 10-fold cross-validation. We have also shown that adding the 23 medication features to the data improves Marco Recall of five out of the six applied learning algorithms. This is a new result that expands the previous studies conducted on the same data.

We believe this work can be further extended and optimized. For example, we found that the major difficulty comes from the instances with "NO" readmission and ">30" readmission days. This requires further investigation. Using Deep Learning algorithms can be a major step in analysis of the data.

# Appendix 1

*23 medication features, with values "Up", "Steady", "Down" and "No"*

Metformin, repaglinide, nateglinide, chlorpropamide, glimepiride, acetohexamide, glipizide, glyburide, tolbutamide, pioglitazone, rosiglitazone, acarbose, miglitol, troglitazone, tolazamide, examide, citoglipton, glyburide-metformin, glipizide-metformin, glimepiride-pioglitazone, metformin-rosiglitazone, metformin-pioglitazone

# Appendix 2

## Naïve Bayes (Gaussian NB)

| Evaluation | Value |
|---|---|
| f-1 score/ precision /recall (micro): | 46.23% |
| Training time | 0.1589 sec |

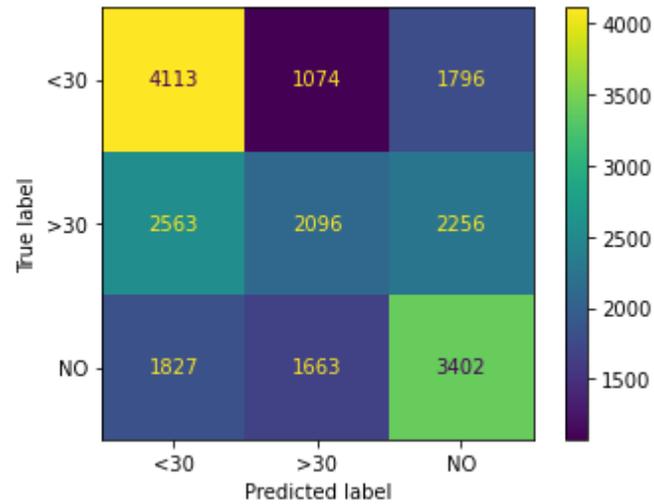

*Confusion matrix of Naïve Bayes*

## SVM

| Evaluation | Value |
|---|---|
| f-1 score/precision/recall (micro) | 53.56% |
| Training time | 21.26 mins |

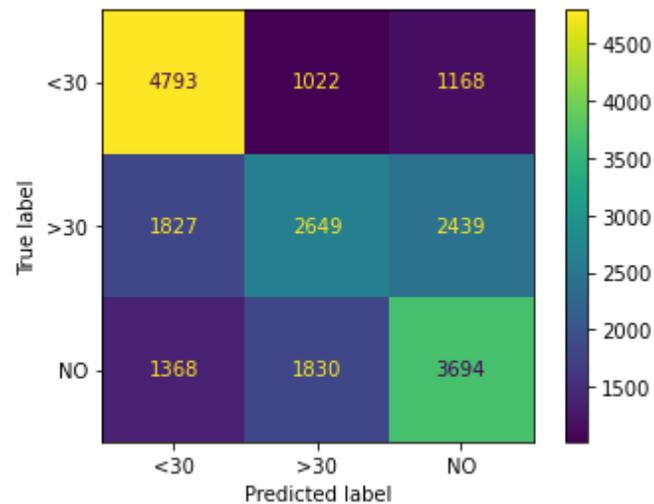

## Gradient Boosting

| Evaluation | Value |
|---|---|
| f-1 score/precision/recall (micro) | 64.04% |
| Training time | 88.81 sec |

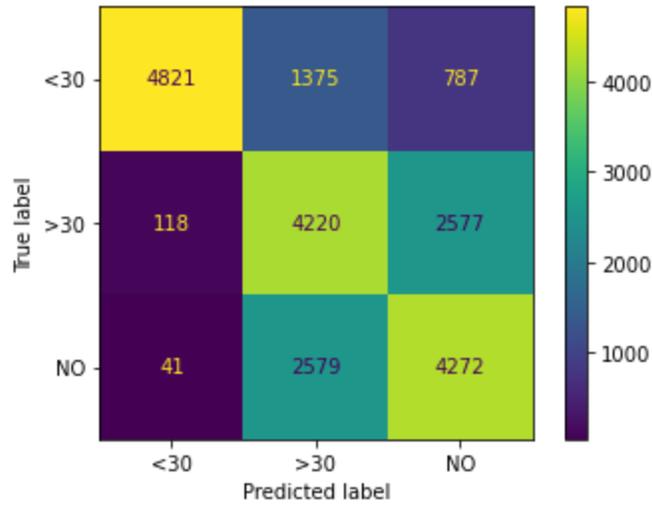

*Confusion matrix of Gradient Boosting*

## Random Forest

| Evaluation | Value |
|---|---|
| f-1 score/precision/recall (micro) | 61.85% |
| Training time | 4.75 sec |

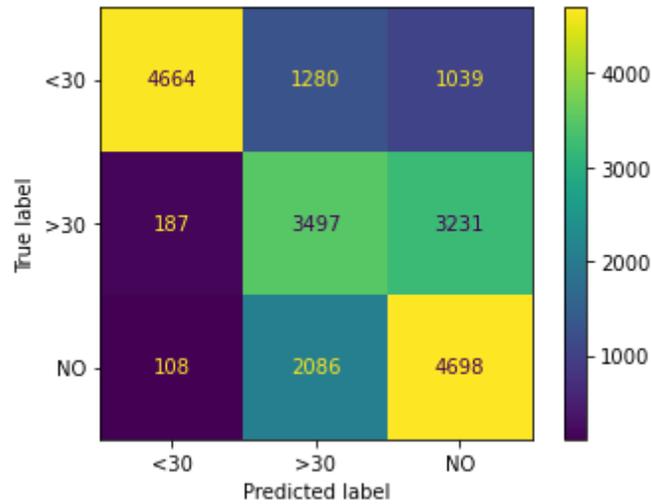

*Confusion matrix of Random Forest*

### Decision Tree

| Evaluation | Value |
|---|---|
| f-1 score/precision/recall (micro) | 59.86% |
| Training time | 0.702 sec |

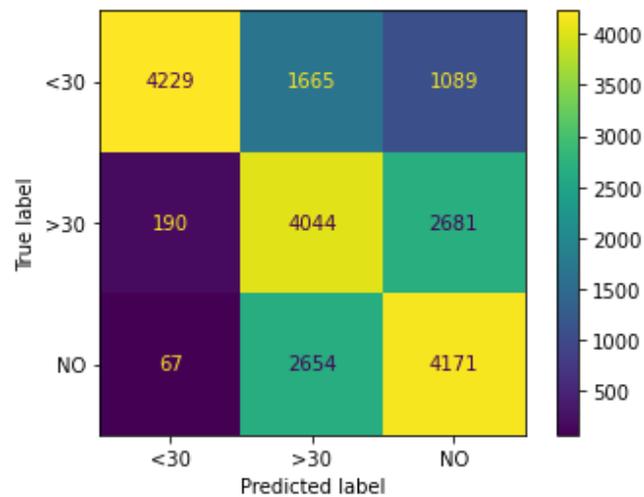

*Confusion matrix of Decision Tree*

### Logistic Regression

| Evaluation | Value |
|---|---|
| f-1 score/precision/recall (micro) | 45.23% |
| Training time | 28.07 sec |

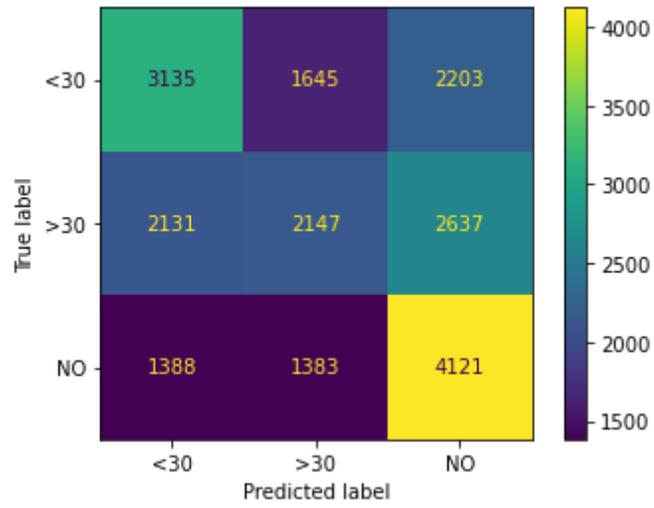

*Confusion matrix of Logistic Regression*